\documentclass{article} 
\usepackage{nips12submit_e,times}

\usepackage{graphicx}
\usepackage{amsmath,amssymb} 
\usepackage{lineno}
\usepackage{color}
\usepackage{algorithm2e}
\usepackage{bm}
\usepackage{subfigure}

\usepackage{times}
\usepackage{epsfig}
\usepackage{amsmath}
\usepackage{amssymb}
\usepackage{bm}

\begin{document}

\title{Complexity of Representation and Inference in Compositional Models with Part Sharing}

\author{%
Alan~L. Yuille\\
Depts. of Statistics, Computer Science \& Psychology\\
University of California, Los Angeles\\
\texttt{yuille@stat.ucla.edu}%
\And
Roozbeh Mottaghi\\
Department of Computer Science\\
University of California, Los Angeles\\
\texttt{roozbehm@cs.ucla.edu}\\
}

\nipsfinalcopy
\maketitle \thispagestyle{empty}

\begin{abstract}

This paper describes serial and parallel compositional models of multiple objects with part sharing. Objects are built by part-subpart compositions and expressed in terms of a hierarchical dictionary of object parts. These parts are represented on lattices of decreasing sizes which yield an executive summary description. We describe inference and learning algorithms for these models.  We analyze the complexity of this model in terms of computation time (for serial computers) and numbers of nodes (e.g., "neurons") for parallel computers. In particular, we compute the complexity gains by part sharing and its dependence on how the dictionary scales with the level of the hierarchy. We explore three regimes of scaling behavior where the dictionary size (i) increases exponentially with the level, (ii)  is determined by an unsupervised compositional learning algorithm applied to real data, (iii) decreases exponentially with scale. This analysis shows that in some regimes the use of shared parts enables algorithms which can perform inference in time linear in the number of levels for an exponential number of objects. In other regimes part sharing has little advantage for serial computers but can give linear processing on parallel computers.


\end{abstract}

\section{Introduction}

A fundamental problem of vision is how to deal with the enormous complexity of images
and visual scenes \footnote{Similar complexity issues will arise for other perceptual and cognitive modalities.}. The total number of possible images is almost infinitely large \cite{Kersten:1987ug}. The number of objects is also huge and has been estimated at around 30,000 \cite{Biederman:1987tc}. How can a biological, or artificial, vision system deal with this complexity? For example, considering the enormous input space of images and output space of objects, how can humans interpret images in less than 150 msec \cite{Thorpe96}? 

There are three main issues involved. Firstly, how can a visual system be designed so that it can efficiently \emph{represent}  large classes of objects, including their parts and subparts? Secondly, how can the visual system be designed so that it can rapidly \emph{infer} which object, or objects, are present in an input image and the positions of their subparts? And, thirdly, how can this representation be \emph{learnt} in an unsupervised, or weakly supervised fashion? In short, what visual \emph{architectures} enable us to address these three issues? 

Many considerations suggest that visual architectures should be hierarchical. The structure of mammalian visual systems is hierarchical with the lower levels (e.g., in areas V1 and V2) tuned to small image features while the higher levels (i.e. in area IT) are tuned to objects \footnote{But just because mammalian visual systems are hierarchical does not necessarily imply that this is the best design for computer vision systems.}. Moreover, as appreciated by pioneers such as Fukushima \cite{FUKUSHIMA:1988wv}, hierarchical architectures lend themselves naturally to efficient representations of objects in terms of parts and subparts which can be shared between many objects. Hierarchical architectures also lead to efficient learning algorithms as illustrated by deep belief learning and others \cite{Hinton06}. There are many varieties of hierarchical models which differ in details of their representations and their learning and inference algorithms \cite{Riesenhuber99, Hinton06, Serre07, Adams03, Poon11, Zeiler10, lecun-bengio-95a, Borenstein:2002uw, KoYu11}. But, to the best of our knowledge, there has been no detailed study of their complexity properties. 

This paper provides a mathematical analysis of compositional models \cite{Geman:2002tg}, which are a subclass of the hierarchical models. The key idea of compositionality is to explicitly represent objects by recursive composition from parts and subparts. This gives rise to natural learning and inference algorithms which proceed from sub-parts to parts to objects (e.g., inference is efficient because a leg detector can be used for detecting the legs of cows, horses, and yaks). The explicitness of the object representations helps quantify the efficiency of part-sharing and make mathematical analysis possible. The compositional models we study are based on the work of L. Zhu and his collaborators \cite{LZhu10unsupervised, zhu08eccv} but we make several technical modifications including a parallel re-formulation of the models. We note that in previous papers \cite{LZhu10unsupervised, zhu08eccv} the representations of the compositional models were learnt in an unsupervised manner, which relates to the memorization algorithms of Valiant \cite{Valiant00}. This paper does not address learning but instead explores the consequence of the representations which were learnt. 

Our analysis assumes that objects are represented by  hierarchical graphical probability models \footnote{These graphical models contain closed loops but with restricted maximal clique size.} which are composed from more elementary models by \emph{part-subpart compositions}. An object -- a graphical model with $\mathcal{H}$ levels -- is defined as a composition of $r$ parts which are graphical models with $\mathcal{H}-1$ levels. These parts are defined recursively in terms of subparts which are represented by graphical models of increasingly lower levels. It is convenient to specify these compositional models in terms of a set of dictionaries $\{\mathcal{M}_h:h=1,..,,\mathcal{H}\}$ where the level-$h$ parts in dictionary $\mathcal{M}_h$ are composed in terms of level-$h-1$ parts in dictionary $\mathcal{M}_{h-1}$. The highest level dictionaries $\mathcal{M}_{\mathcal{H}}$ represent the set of all objects. The lowest level dictionaries $\mathcal{M}_1$ represent the elementary features that can be measured from the input image. Part-subpart composition enables us to construct a very large number of objects by different compositions of elements from the lowest-level dictionary. It enables us to perform \emph{part-sharing} during learning and inference, which can lead to enormous reductions in complexity, as our mathematical analysis will show.

There are three factors which enable computational efficiency. The first is \emph{part-sharing}, as described above, which means that we only need to perform inference on the dictionary elements. The second is the \emph{executive-summary principle}. This principle allows us to represent the state of a part coarsely because we are also representing the state of its subparts (e.g., an executive will only want to know that "there is a horse in the field" and will not care about the precise positions of its legs). For example, consider a letter $T$ which is composed of a horizontal and vertical bar. If the positions of these two bars are specified precisely, then we can specify the position of the letter $T$ more crudely (sufficient for it to be "bound" to the two bars). This relates to Lee and Mumford's high-resolution buffer hypothesis \cite{LeeMumford2003} and possibly to the experimental finding that neurons higher up the visual pathway are tuned to increasingly complex image features but are decreasingly sensitive to spatial position. The third factor is \emph{parallelism} which arises because the part dictionaries can be implemented in parallel, essentially having a set of receptive fields, for each dictionary element. This enables extremely rapid inference at the cost of a larger, but parallel, graphical model.

The compositional section~(\ref{sec:compositional}) introduces the key ideas. Section~(\ref{sec:inference}) describes the inference algorithms for serial and parallel implementations. Section~(\ref{sec:complexity}) performs a complexity analysis and shows potential exponential gains by using compositional models. 


\section{The Compositional Models \label{sec:compositional}}

Compositional models are based on the idea that objects are built by compositions of parts which, in turn, are compositions of more elementary parts. These are built by part-subpart compositions.

\subsection{Compositional Part-Subparts}

We formulate part-subpart compositions by probabilistic graphical model which specifies how a part is composed of its subparts. A parent node $\nu$ of the graph represents the part by its type $\tau _{\nu}$ and a state variable $x_{\nu}$ (e.g., $x_{\nu}$ could indicate the position of the part). The $r$ child nodes $Ch(\nu) = (\nu _1,...,\nu _r)$ represent the parts by their types $\tau _{\nu _1},...,\tau _{\nu _r}$ and state variables $\vec x _{Ch(\nu)}= (x_{\nu _1},...,x_{\nu _r})$ \footnote{In this paper, we assume a fixed value $r$ for all part-subpart compositions.}. The type of the parent node is specified by $\tau _{\nu} = (\tau _{\nu _1},...,\tau _{\nu _r}, \lambda _{\nu})$. Here $(\tau _{\nu _1},...,\tau _{\nu _r})$ are the types of the child nodes, and $\lambda _{\nu}$ specifies a distribution over the states of the subparts (e.g.,over their relative spatial positions). Hence the type $\tau _{\nu}$ of the parent specifies the part-subpart compositional model.

The probability distribution for the part-subpart model relates the states of the part and the subparts  by:
\begin{equation} P(\vec x _{Ch(\nu)}|x _{\nu}; \tau _{\nu}) = \delta (x_{\nu} - f (\vec x _{Ch(\nu)})) h(\vec x _{Ch(\nu)};\lambda _{\nu}).\label{parent_child}\end{equation}
Here $f(.)$ is a deterministic function, so the state of the parent node is determined uniquely by the state of the child nodes. The function $h(.)$ specifies a distribution on the relative states of the child nodes. The distribution $P(\vec x _{Ch(\nu)}|x _{\nu}; \tau _{\nu})$ obeys a \emph{locality principle}, which means that $P(\vec x _{Ch(\nu)}|x _{\nu}; \tau _{\nu}) =0$, unless $|x _{\nu _i} - x _{\nu}|$ is smaller than a threshold for all $i=1,..,r$. This requirement captures the intuition that subparts of a part are typically close together.

The state variable $x_{\nu}$ of the parent node provides an \emph{executive summary} description of the part. Hence they are restricted to take a smaller set of values than the state variables $x_{\nu _i}$ of the subparts. Intuitively, the state $x_{\nu}$ of the parent offers summary information (e.g., there is a cow in the right side of a field) while the child states $\vec x _{Ch(\nu)}$ offer more detailed information (e.g., the position of the parts of the cow). In general, information about the object is represented in a distributed manner with coarse information at the upper levels of the hierarchy and more precise information at lower levels.

\begin{figure}[tp]
\centering
\subfigure[]{
   \includegraphics[width=17pc] {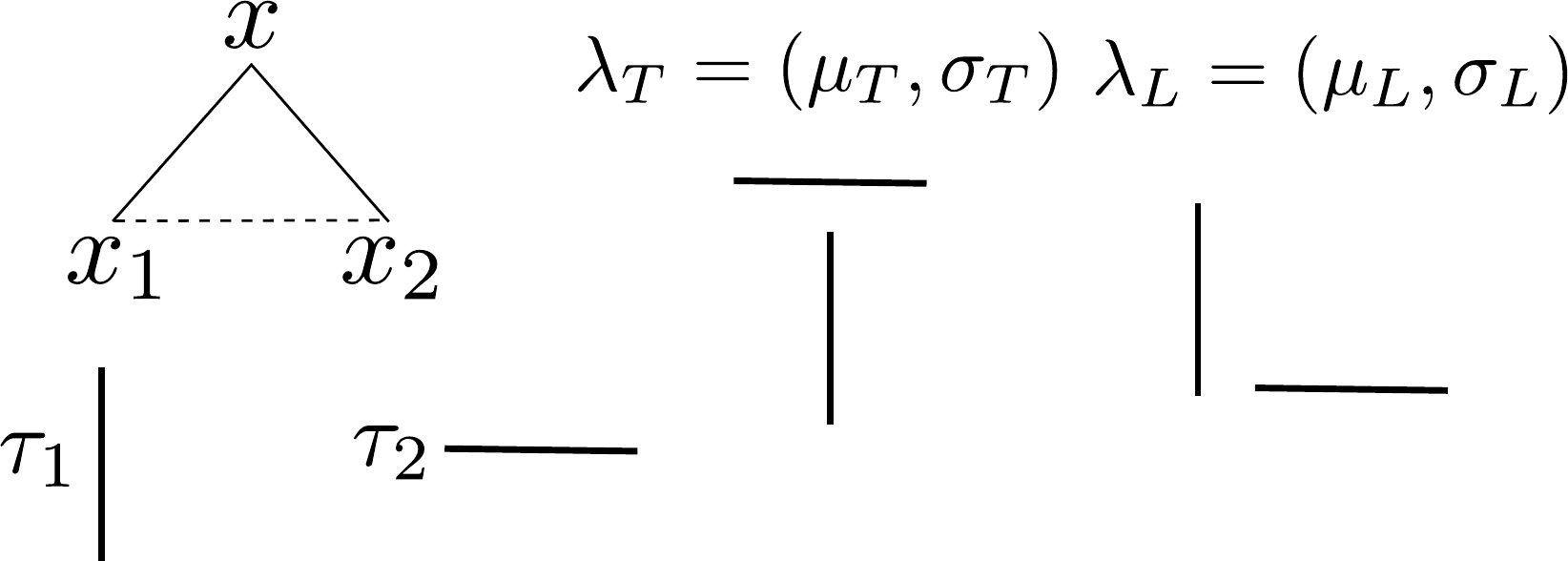}
   \label{fig:comp1}
 }
 \subfigure[]{
   \includegraphics[width=10pc] {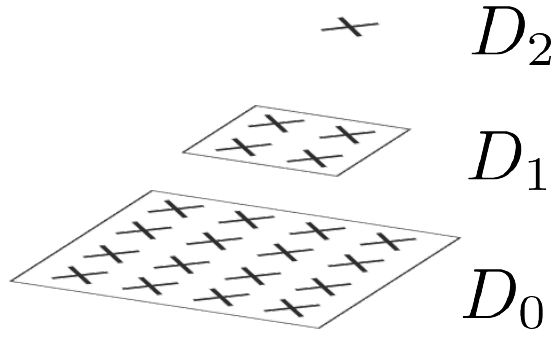}
   \label{fig:lattices}
 }
\label{fig:n}
\caption{(a) Compositional part-subpart models for $T$ and $L$ are constructed from the same elementary components $\tau _1, \tau _2$, horizontal and vertical bar using  different spatial relations $\lambda=(\mu, \sigma)$, which impose {\it locality}. The state $x$ of the parent node gives the summary position of the object, the \emph{executive summary}, while the positions $x_1,x_2$ of the components give details about its components. (b) The hierarchical lattices. The size of the lattices decrease with scale by a factor $q$ which helps enforce executive summary and prevent having multiple hypotheses which overlap too much. $q= 1/4$ in this figure.}
\end{figure}

We give examples of part-subpart compositions in figure~(\ref{fig:comp1}). The compositions represent the letters $T$ and $L$, which are the types of the parent nodes. The types of the child nodes are horizontal and vertical bars, indicated by $\tau _1 = H, \ \tau _2 = V$. The child state variables $x_1,x_2$ indicate the image positions of the horizontal and vertical bars. The state variable $x$ of the parent node gives a summary description of the position of the letters $T$ and $L$.  The compositional models for letters $T$ and $L$ differ by their $\lambda$ parameter which species the relative positions of the horizontal and vertical bars. In this example, we choose $h(.;\lambda)$ to a Gaussian distribution, so $\lambda = (\mu,\sigma)$ where $\mu$ is the mean relative positions between the bars and $\sigma$ is the covariance. We set $f(x_1,x_2) = (1/2) (x_1+x_2)$, so the state of the parent node specifies the average positions of the child nodes (i.e. the positions of the two bars). Hence the two compositional models for the $T$ and $L$ have types $\tau _T = (H,V,\lambda _T)$ and $\tau _L = (H,V, \lambda _L)$.


\subsection{Models of Object Categories}

\begin{figure}[tp]
\centering
\includegraphics[width=28pc]{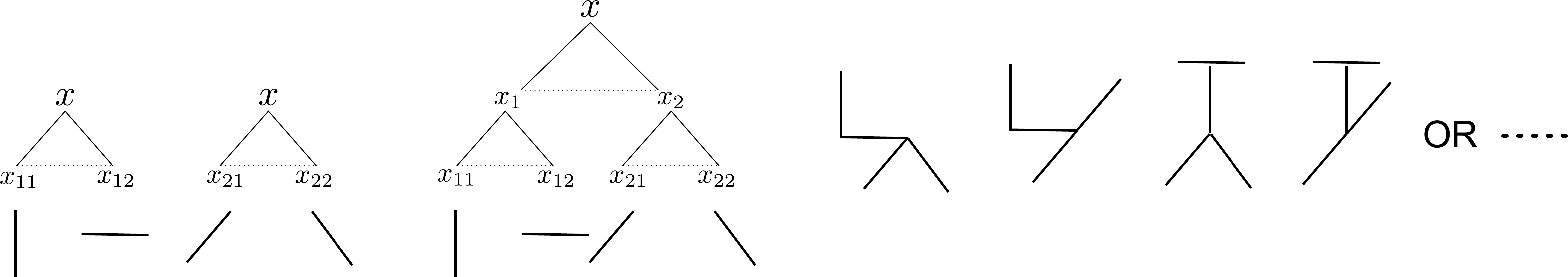}
\caption{Left Panel: Two part-subpart models. Center Panel: Combining two part-subpart models by composition to make a higher level model. Right Panel: Some examples of the shapes that can be generated by different parameters settings $\lambda$ of the distribution.} \label{fig:comp2}
\end{figure}

An object category can be modeled by repeated part-subpart compositions. This is illustrated in figure~(\ref{fig:comp2}) where we combine $T$'s and $L$'s with other parts to form more complex objects. More generally, we can combine part-subpart compositions into bigger structures by treating the parts as subparts of higher order parts.

More formally, an object category of type $\tau _{\mathcal{H}}$ is represented by a probability distribution defined over a graph $\mathcal{V}$. This graph has a hierarchical structure with levels $h
\in \{0,...,\mathcal{H}\}$, where $\mathcal{V} = \bigcup _{h=0}^{\mathcal{H}} \mathcal{V}_h$. Each object has
a single, root node, at level-$\mathcal{H}$  (i.e. $\mathcal{V}_{\mathcal{H}}$ contains a single node). Any node
$\nu \in \mathcal{V}_h$ (for $h>0$) has $r$ children nodes $Ch(\nu)$ in
$\mathcal{V}_{h-1}$ indexed by $(\nu _1,...,\nu _r)$. Hence there are $r ^{\mathcal{H}-h}$ nodes
at level-h (i.e. $|\mathcal{V}_h|= r ^{\mathcal{H} -h}$). 

At each node $\nu$ there is a state variable $x_{\nu}$ which indicates spatial position
and type $\tau _{\nu}$. The type $\tau _{\mathcal{H}}$ of the root node indicates the object category and also specifies the types of its parts. 

The position variables $x_{\nu}$ take values in a \emph{set of lattices} $\{\mathcal{D}_h: h=0,...,\mathcal{H}\}$, so that a level-h node, $\nu \in \mathcal{V}_h$, takes position  $x_{\nu} \in \mathcal{D}_h$. The leaf nodes $\mathcal{V}_0$ of the graph take values on the image lattice $\mathcal{D}_0$. The lattices are evenly spaced and the number of lattice points decreases by a factor of $q<1$ for each level, so $|\mathcal{D}_h| = q ^h|\mathcal{D}_0|$, see figure~(\ref{fig:lattices}). This decrease in number of lattice points imposes the \emph{executive summary principle}. The lattice spacing is designed so that parts do not overlap. At higher levels of the hierarchy the parts cover larger regions of the image and so the lattice spacing must be larger, and hence the number of lattice points smaller, to prevent overlapping \footnote{Previous work \cite{zhu08eccv,LZhu10unsupervised} was not formulated on lattices and used non-maximal suppression to achieve the same effect.}.

%
%
The probability model for an object category of type $\tau _{\mathcal{H}}$ is specified by products of part-subpart relations:
\begin{equation} P(\vec x| \tau _{\mathcal{H}}) =  \prod _{\nu \in \mathcal{V}/\mathcal{V}_0} P(\vec x _{Ch(\nu})|x_{\nu};\tau_{\nu}) U(x_{\mathcal{H}}).\label{eq:prior}\end{equation}
Here $U(x_{\mathcal{H}})$ is the uniform distribution. 


\subsection{Multiple Object Categories, Shared Parts, and  Hierarchical Dictionaries}

Now suppose we have a set of object categories $\tau _{\mathcal{H}} \in \mathcal{H}$, each of which can be expressed by an equation such as equation~(\ref{eq:prior}). We assume that these objects  share parts. To quantify the amount of part sharing we define a hierarchical dictionary $\{\mathcal{M}_h:h=0,...,{\mathcal{H}}\}$, where $\mathcal{M}_h$ is the dictionary of parts at level $h$. This gives an exhaustive set of the parts of this set of the objects, at all levels $h=0,...,\mathcal{H}$. The elements of the dictionary $\mathcal{M}_h$ are composed from elements of the dictionary $\mathcal{M}_{h-1}$ by part-subpart compositions \footnote{The unsupervised learning algorithm in \cite{LZhu10unsupervised} automatically generates this hierarchical dictionary.}.

This gives an alternative way to think of object models. The type variable $\tau _{\nu}$ of a node at level $h$ (i.e. in $\mathcal{V}_h$) indexes an element of the dictionary $\mathcal{M}_h$. Hence objects can be encoded in terms of the hierarchical dictionary. Moreover, we can create new objects by making new compositions from existing elements of the dictionaries. 


\subsection{The Likelihood Function and the Generative Model}

To specify a generative model for each object category we proceed as follows. The prior specifies a distribution over the positions and types of the leaf nodes of the object model. Then the likelihood function is specified in terms of the type at the leaf nodes (e.g., if the leaf node is a vertical bar, then there is a high probability that the image has a vertical edge at that position).

More formally, the prior $P(\vec x| \tau _{\mathcal{H}})$, see equation~(\ref{eq:prior}), specifies a distribution over a set of points
$\mathcal{L} = \{ x_{\nu}: \nu \in \mathcal{V}_0\}$ (the leaf nodes of the graph) and specifies their types $\{\tau _{\nu}: \nu \in \mathcal{V}_0\}$. These points are required to lie on the image lattice (e.g., $x_{\nu} \in \mathcal{D}_0$). We denote this as $\{(x,\tau(x)): x \in \mathcal{L}\}$ where $\tau(x)$ is specified in the natural manner (i.e. if $x = x_{\nu}$ then $\tau (x) = \tau _{\nu}$). We specify distributions $P(I(x)|\tau(x))$ for the probability of the image $I(x)$ at $x$ conditioned on the type of the leaf node. We specify a default probability $P(I(x)|\tau _0)$ at positions $x$ where there is no leaf node of the object.

This gives a likelihood function for the states $\vec x = \{x _{\nu} \in \mathcal{V}\}$ of the object model in terms of the image $\bf I = \{I(x): x \in \mathcal{D}_0\}$:
\begin{equation}P({\bf I}|\vec x) =
\prod _{x \in \mathcal{L}} P(I(x)|\tau (x)) \times \prod _{x \in
\mathcal{D}_0/\mathcal{L}} P(I(x)|\tau _0).\label{eq:likelihood}\end{equation}
The likelihood and the prior, equations~(\ref{eq:likelihood},\ref{eq:prior}), give a generative model for each object category. 

We can extend this in the natural manner to give generative models or two, or more, objects in the image provided they do not overlap. Intuitively, this involves multiple sampling from the prior to determine the types of the lattice pixels, followed by sampling from $P(I(x)|\tau)$ at the leaf nodes to determine the image $\bf I$. Similarly, we have a default \emph{background model} for the entire image if no object is present:
\begin{equation} P_B({\bf I}) = \prod _{x \in \mathcal{D}_0} P(I(x)|\tau _0).\label{eq:background}\end{equation}

\section{Inference by Dynamic Programming \label{sec:inference}}

The inference task is to determine which objects are present in the image and to specify their positions. This involves two subtasks: (i) state estimation, to determine the optimal states of a model and hence the position of the objects and its parts, and (ii) model selection, to determine whether objects are present or not. As we will show, both tasks can be reduced to calculating and comparing log-likelihood ratios which can be performed efficiently using dynamic programming methods.

We will first describe the simplest case which consists of estimating the state variables of a single object model and using model selection to determine whether the object is present in the image and, if so, how many times. Next we show that we can  perform inference and model selection for multiple objects efficiently by exploiting part sharing (using hierarchical dictionaries). Finally, we show how these inference tasks can be performed even more efficiently using a parallel implementation. We stress that we are performing \emph{exact inference} and no approximations are made. We are simply exploiting part-sharing so that computations required for performing inference for one object can be re-used when performing inference for other objects.

\subsection{Inference Tasks: State Detection and Model Selection}

We first describe a standard dynamic programming algorithm for finding the optimal state of a single object category model. Then we describe how the same computations can be used to perform model selection and to the detection and state estimation if the object appears multiple times in the image (non-overlapping).

Consider performing inference for a single object category model defined by equations~(\ref{eq:prior},\ref{eq:likelihood}). To calculate the MAP estimate of the state variables requires computing $\vec x^{\ast} = \arg \max _{\vec x} \{ \log P({\bf I}|\vec x) + \log P(\vec x; \tau _{\mathcal{H}})\}$. By subtracting the constant term $\log P_B({\bf I})$ from the righthand side, we can re-express this as estimating:

\begin{eqnarray}  \vec x^{\ast} = \arg \max _{\vec x}  \{ \sum _{ x \in \mathcal{L}}  \log {{
P(I(x)|\tau(x))}\over{P(I(x)|\tau _0)}} + \sum _{\nu} \log P(\vec
x_{Ch(\nu)}|x_{\nu};\tau _{\nu}) + \log U(x_{\mathcal{H}})\}.\label{eq:ML1}\end{eqnarray}  
Here $\mathcal{L}$ denotes the positions of the leaf nodes of the graph, which must be determined during inference.

We estimate $\vec x^{\ast}$ by performing dynamic programming. This involves a bottom-up pass which recursively computes quantities $\phi (x_h, \tau _h)=
\arg \max _{\vec x/x_h} \{ \log {{P({\bf I}|\vec x)}\over{P_B({\bf I})}} + \log P(\vec x; \tau _h)\}$ by the formula:
\begin{equation}\phi (x_h, \tau _h) = \max _{\vec x_{Ch(\nu)}}
\{ \sum _{i=1}^r \phi (x_{\nu _i},\tau _{\nu _i}) + \log P(\vec x_{Ch(\nu)}| x ^{\ast}_{\nu},\tau
_{\nu})\}. \label{eq:bottom_up}\end{equation}
We refer to $\phi (x_h, \tau _h)$ as the \emph{local evidence} for part $\tau _h$ with state $x_h$ (after maximizing over the states of the lower parts of the graphical model). This local evidence is computed bottom-up. We call this the local evidence because it ignores the context evidence for the part which will be provided during top-down processing (i.e. that evidence for other parts of the object, in consistent positions, will strengthen the evidence for this part).

\begin{figure}[tp]
\centering
\includegraphics[width=28pc]{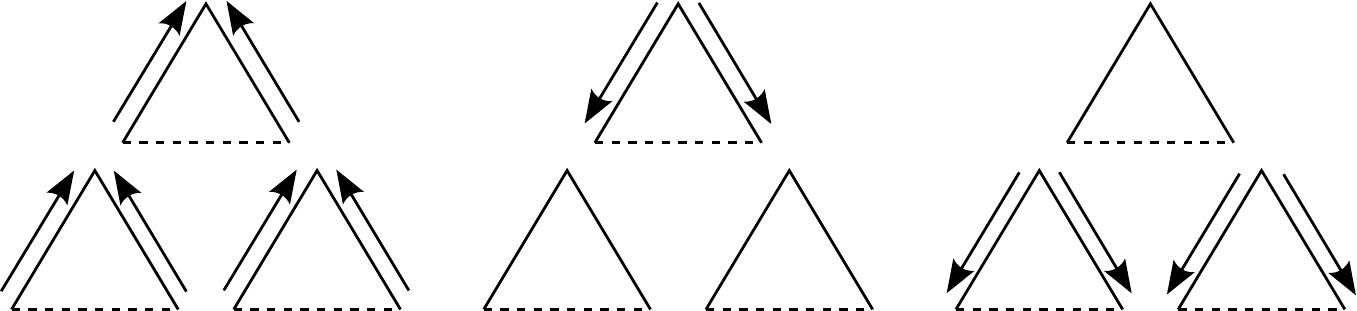}
\caption{Left Panel: The feedforward pass propagates hypotheses up to the highest level where the best
state is selected. Center Panel: Feedback propagates information from the top node
disambiguating the middle level nodes. Right Panel: Feedback from the middle level nodes
propagates back to the input layer to resolve ambiguities there. This algorithm rapidly estimates the top-level executive summary description in a rapid feed-forward pass. The top-down pass is required to allow high-level context to eliminate false hypotheses at the lower levels-- "high-level tells low-level to stop gossiping".} \label{fig:comp3}
\end{figure}

The bottom-up pass outputs the \emph{global evidence} $\phi(x_{\mathcal{H}},\tau_{\mathcal{H}})$ for object category $\tau _{\mathcal{H}}$ at position $x_{\mathcal{H}}$. We can detect the most probable state of the object by computing $x_{\mathcal{H}}^{\ast} = \arg \max \phi(x_{\mathcal{H}},\tau_{\mathcal{H}})$. Then we can perform the top-down pass of dynamic programming to estimate the most probable states $\vec x^{\ast}$ of the entire model by recursively performing:
\begin{equation} \vec x ^{\ast}_{Ch(\nu)}= \arg \max _{\vec x_{Ch(\nu)}}
\{ \sum _{i=1}^r \phi (x_{\nu _i},\tau _{\nu _i}) + \log P(\vec x_{Ch(\nu)}| x ^{\ast}_{\nu},\tau
_{\nu})\}.\label{eq:top_down}\end{equation}
This outputs the most probable state of the object in the image. Note that the bottom-up process first estimates the optimal "executive summary" description of the object ($x_{\mathcal{H}}^{\ast}$) and only later determines the optimal estimates of the lower-level states of the object in the top-down pass. Hence, the algorithm is faster at detecting that there is a cow in the right side of the field (estimated in the bottom-up pass) and is slower at determining the position of the feet of the cow (estimated in the top-down pass). This is illustrated in figure~(\ref{fig:comp3}).

Importantly, we only need to perform slight extensions of this algorithm to compute significantly more. First, we can perform model selection --  to determine if the object is present in the image -- by determining if 
$\phi(x_{\mathcal{H}}^{\ast},\tau _{\mathcal{H}}) > T$, where $T$ is a threshold. This is because, by equation~(\ref{eq:ML1}), $\phi(x_{\mathcal{H}}^{\ast},\tau _{\mathcal{H}})$ is the log-likelihood ratio of the probability that the object is present at position $x_{\mathcal{H}}^{\ast}$ compared to the probability that the corresponding part of the image is generated by the background image model $P_B(.)$. Secondly, we can compute the probability that the object occurs several times in the image, by computing the set $\{ x_{\mathcal{H}}:
\phi(x_{\mathcal{H}}^{\ast},\tau _{\mathcal{H}}) > T$, to compute the "executive summary" descriptions for each object (e.g., the coarse positions of each object). We then perform the top-down pass initialized at each coarse position (i.e. at each point of the set described above) to determine the optimal configuration for the states of the objects. Hence, we can reuse the computations required to detect a single object in order to detect multiple instances of the object (provided there are no overlaps)\footnote{Note that this is equivalent to performing optimal inference simultaneously over a set of different generative models of the image, where one model assumes that there is one instance of the object in the image, another models assumes there are two, and so on.}. The number of objects in the image is determined by the log-likelihood ratio test with respect to the background model.

\subsection{Inference on Multiple Objects by Part Sharing using the Hierarchical Dictionaries}

Now suppose we want to detect instances of many object categories $\tau _{\mathcal{H}} \in \mathcal{M}_{\mathcal{H}}$ simultaneously. We can  exploit the shared parts by performing inference using the hierarchical dictionaries.

The main idea is that we need to compute the global evidence $\phi (x_{\mathcal{H}},\tau_{\mathcal{H}})$ for all objects $\tau _{\mathcal{H}} \in \mathcal{M}_{\mathcal{H}}$ and at all positions $x_{\mathcal{H}}$ in the top-level lattice. These quantities could be computed separately for each object by performing the bottom-up pass, specified by equation~(\ref{eq:bottom_up}), for each object. But this is wasteful because the objects share parts and so we would be performing the same computations multiple times. Instead we can perform all the necessary computations more efficiently by working directly with the hierarchical dictionaries.

More formally, computing the global evidence for all object models and at all positions is specified as follows.
\begin{eqnarray} {\rm Let} \  \mathcal{D}_{\mathcal{H}}^{\ast} = \{x _{\mathcal{H}} \in \mathcal{D}_{\mathcal{H}} \ {\rm s.t.} \ \max _{\tau _{\mathcal{H}} \in \mathcal{M}_{\mathcal{H}}} \phi(x_{\mathcal{H}},\tau _{\mathcal{H}}) > T_{\mathcal{H}} \}, \nonumber \\
{\rm For} \ x_{\mathcal{H}} \in \mathcal{D}_{\mathcal{H}}^{\ast}, \ {\rm let} \ \tau _{\mathcal{H}}^{\ast}(x_{\mathcal{H}}) = \arg \max _{\tau _{\mathcal{H}} \in \mathcal{M}_{\mathcal{H}}} \phi(x_{\mathcal{H}},\tau _{\mathcal{H}}),\nonumber \\
{\rm Detect} \ \vec x ^{\ast}/x_{\mathcal{H}} = \arg \max _{\vec x /x_{\mathcal{H}}}  \{ \log {{P({\bf I}|\vec x)}\over{P_B({\bf I})}} + \log P(\vec x; \tau _{{\mathcal{H}^{\ast}}(x_{\mathcal{H}})})\} \ {\rm for \ all} \ x_{\mathcal{H}} \in \mathcal{D}_{\mathcal{H}}^{\ast}.\label{eq:fourthtask}\end{eqnarray}

\begin{figure}[tp]
\centering
\includegraphics[width=28pc]{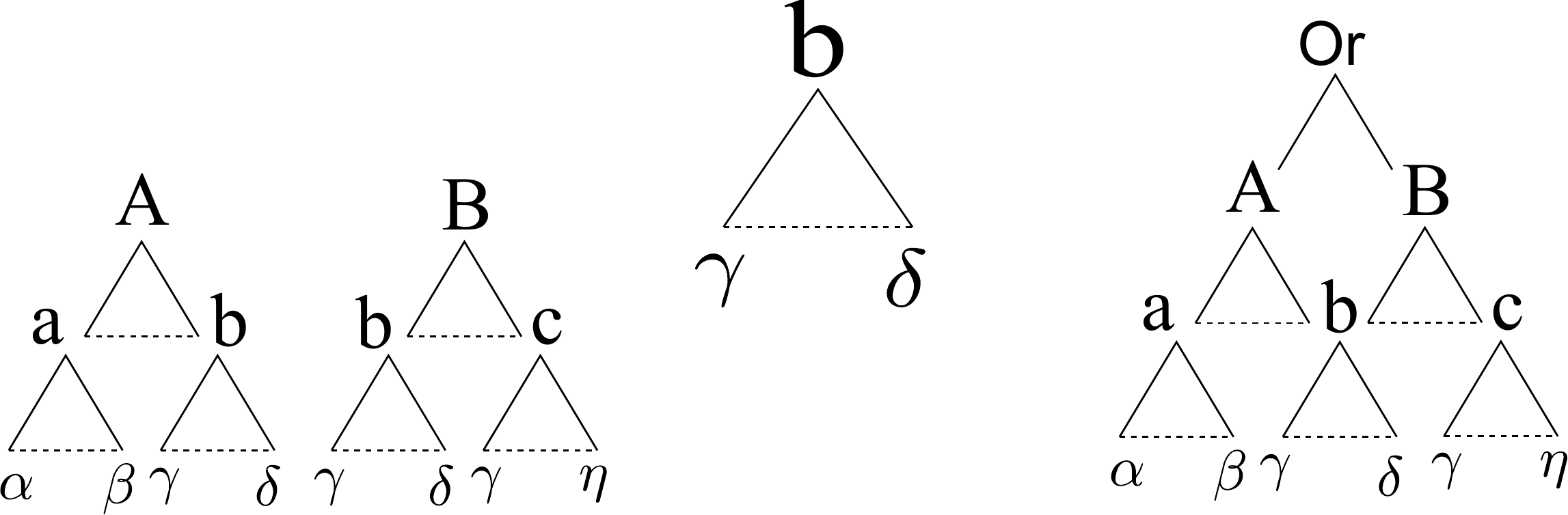}
\caption{Sharing. Left Panel: Two Level-2 models $A$ and $B$ which share Level-1 model $b$ as a subpart. Center Panel:
Level-1 model $b$. Inference computation only requires us to do inference over model $b$
once, and then it can be used to computer the optimal states for models $A$ and $B$.
Right Panel: Note that is we combine models $A$ and $B$ by a root OR node then we obtain
a graphical model for both objects. This model has a closed loop which would seem to
make inference more challenging. But by exploiting the shape part we can do inference
optimally despite the closed loop. Inference can be done on the dictionaries, far right.}\label{fig:comp4}
\end{figure}
 
All these calculations can be done efficiently using the hierarchical dictionaries (except for the $\max$ and $\arg \max$ tasks at level $\mathcal{H}$ which must be done separately). Recalling that each dictionary element at level $h$ is composed, by part-subpart composition, of dictionary elements at level $h-1$. Hence we can apply the bottom-up update rule in equation~(\ref{eq:bottom_up}) directly to the dictionary elements. This is illustrated in figure~(\ref{fig:comp4}). As analyzed in the next section, this can yield major gains in computational complexity.

Once the global evidences for each object model have been computed at each position (in the top lattice) we can perform winner-take-all to estimate the object model which has largest evidence at each position. Then we can apply thresholding to see if it passes the log-likelihood ratio test compared to the background model. If it does pass this log-likelihood test, then we can use the top-down pass of dynamic programming, see equation~(\ref{eq:top_down}), to estimate the most probable state of all parts of the object model.

We note that we are performing exact inference over multiple object models at the same time. This is perhaps 
un-intuitive to some readers because this corresponds to doing exact inference over a probability model which can be expressed as a graph with a large number of closed loops, see figure~(\ref{fig:comp4}). But the main point is that part-sharing enables us share inference efficiently between many models.

The only computation which cannot be performed by dynamic programming are the $\max$ and $\arg \max$ tasks at level  $H$, see top line of equation~(\ref{eq:fourthtask}). These are simple operations and require order $M_{\mathcal{H}} \times |\mathcal{D}_{\mathcal{H}}|$ calculations. This will usually be a small number, compared to the complexity of other computations. But this will become very large if there are a large number of objects, as we will discuss in section~(\ref{sec:complexity}).

\subsection{Parallel Formulation and Inference Algorithm}

Finally, we observe that all the computations required for performing inference on multiple objects can be parallelized. This requires computing the quantities in equation~(\ref{eq:fourthtask}).

\begin{figure}[htbp]
\centerline{\psfig{figure=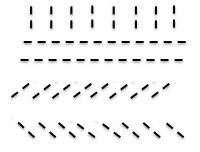,width=0.2\columnwidth}
\psfig{figure=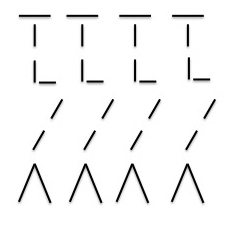,width=0.2\columnwidth}
\psfig{figure=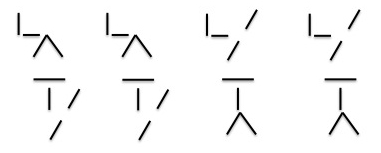,width=0.2\columnwidth}
\psfig{figure=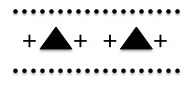,width=0.2\columnwidth}} \caption{Parallel Hierarchical
Implementation. Far Left Panel: Four Level-0 models are spaced densely in the image
(here an $8 \times 2$ grid). Left Panel: the four Level-1 models are sampled at a lower rate,
and each have $4 \times 1$ copies. Right Panel: the four Level-2 models are sampled less
frequently. Far Right Panel: a bird's eye view of the parallel hierarchy. The dots
represent a "column" of four Level-0 models. The crosses represent columns containing
four Level-1 models. The triangles represent a column of the Level-2
models.\label{fig:comp5}}
\end{figure}

 The parallelization is possible, in the bottom-up pass of dynamic programming, calculations are done separately for each position $x$, see equation~(\ref{eq:bottom_up}). So we can compute the local evidence for all parts in the hierarchical dictionary recursively and in parallel for each position, and hence compute the $\phi (x_{\mathcal{H}},\tau _{\mathcal{H}})$ for all $x_{\mathcal{H}} \in \mathcal{D}_{\mathcal{H}}$ and $\tau _{\mathcal{H}} \in \mathcal{M}_{\mathcal{H}}$.
The $\max$ and $\arg \max$ operations at level $\mathcal{H}$ can also be done in parallel for each  position $x_{\mathcal{H}} \in \mathcal{D}_{\mathcal{H}}$. Similarly we can perform the top-down pass of dynamic programming, see equation~(\ref{eq:top_down}), in parallel to compute the best configurations of the detected objects in parallel for different possible positions of the objects (on the top-level lattice).

The parallel formulation can be visualized by making copies of the elements of the hierarchical dictionary elements (the parts), so that a model at level-h has $|\mathcal{D}_h|$ copies, with one copy at each lattice point. Hence at level-h, we have $m_h$ "receptive fields" at each lattice point in $\mathcal{D}_h$ with each one tuned to a different part $\tau _h \in \mathcal{M}_h$, see figure~(\ref{fig:comp5}). At level-0, these receptive fields are tuned to specific image properties (e.g., horizontal or vertical bars). Note that the receptive fields are highly non-linear (i.e. they do not obey any superposition principle)\footnote{Nevertheless they are broadly speaking, tuned to image stimuli which have the mean shape of the corresponding part $\tau _h$. In agreement, with findings about mammalian cortex, the receptive fields become more sensitive to image structure (e.g., from bars, to more complex shapes) at increasing levels. Moreover, their sensitivity to spatial position decreases because at higher levels the models only encode the executive summary descriptions, on coarser lattices, while the finer details of the object are represented more precisely at the lower levels.}. Moreover, they are influenced both by bottom-up processing (during the bottom-up pass) and by top-down processing (during the top-down pass). The bottom-up processing computes the local evidence while the top-down pass modifies it by the high-level context.

\begin{figure}[tp]
\centering
\includegraphics[width=28pc]{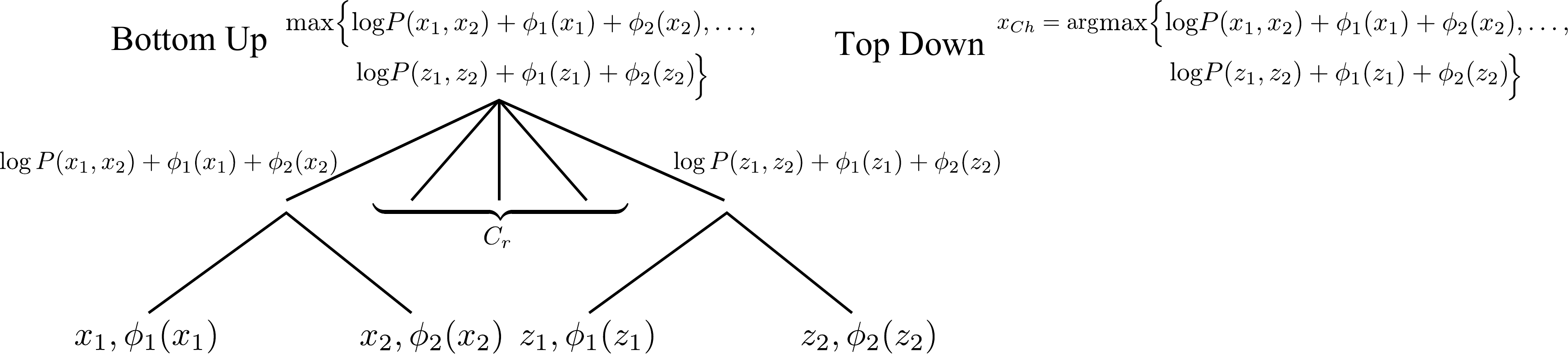}
\caption{Parallel implementation of Dynamic Programming. The left part of the figure shows the bottom-up pass of dynamic programming. The local evidence for the parent node is obtained by taking the maximum of the scores of the $C_r$ possible states of the child nodes. This can be computed by a two-layer network where the first level computes the scores for all $C_r$ child node states, which can be done in parallel, and the second level compute the maximum score. This is like an AND operation followed by an OR. The top-down pass requires the parent node to select which of the $C_r$ child configurations gave the maximum score, and suppressing the other configurations. \label{fig:neural}}
\end{figure}

The computations required by this parallel implementation are illustrated in figure~(\ref{fig:neural}). The bottom-up pass is performed by a two-layer network where the first layer performs an AND operation (to compute the local evidence for a specific configuration of the child nodes) and the second layer performs an OR, or $\max$ operation, to determine the local evidence (by max-ing over the possible child configurations)\footnote{Note that other hierarchical models, including bio-inspired ones, use similar operations but motivated by different reasons.}. The top-down pass only has to perform an $\arg \max$ computation to determine which child configuration gave the best local evidence.

\section{Complexity Analysis \label{sec:complexity}}

We now analyze the complexity of the inference algorithms for performing the tasks. Firstly, we analyze complexity for a single object (without part-sharing). Secondly, we study the complexity for multiple objects with shared parts. Thirdly, we consider the complexity of the parallel implementation.

The complexity is expressed in terms of the following quantities: (I) The size $|\mathcal{D}_0|$ of the image. (II) The scale decrease factor $q$ (enabling executive summary). (III) The number $\mathcal{H}$ of levels of the hierarchy. (IV) The sizes $\{|\mathcal{M}_h|: h=1,...,\mathcal{H}\}$ of the hierarchical dictionaries. (V) The number $r$ of subparts of each part. (VI) The number $C_r$ of possible part-subpart configurations.

\subsection{Complexity for Single Objects and Ignoring Part Sharing}

This section estimates the complexity of inference $N_{s_o}$ for a single object and the complexity $N_{m_o}$ for multiple objects when part sharing is not used. These results are for comparison to the complexities derived in the following section using part sharing.

The inference complexity for a single object requires computing: (i) the number $N_{bu}$ of computations required by the bottom-up pass, (ii) the number $N_{m_s}$  of computations required by  model selection at the top-level of the hierarchy, and (iii) the number $N_{td}$ of computations required by the top-down pass.

The complexity $N_{bu}$ of the bottom-up pass can be computed from equation~(\ref{eq:bottom_up}). This requires a total of $C_r$ computations for each position $x_{\nu}$ for each level-h node. There are $r^{\mathcal{H}-h}$ nodes at level $h$ and each can take $|\mathcal{D}_0| q ^h$ positions. This gives a total of $|\mathcal{D}_0| C_r q ^h r ^{\mathcal{H}-h}$ computations at level $h$. This can be summed over all levels to yield:
\begin{equation} N_{bu} =
\sum _{h=1}^{\mathcal{H}} |\mathcal{D}_0| C _r r ^{\mathcal{H}} (q/r)^h = |\mathcal{D}_0| C _r r ^{\mathcal{H}} \sum _{h=1}^{\mathcal{H}} (q/r)^h =  |\mathcal{D}_0| C _r {{q
r^{\mathcal{H}-1}}\over{1 - q/r}} \{1 - (q/r)^{\mathcal{H}}\}.\label{eq:b_u}\end{equation}
Observe that the main contributions to $N_{bu}$ come from the first few levels of the hierarchy because the factors $(q/r)^h$ decrease rapidly with $h$. This calculation uses $\sum _{h=1}^{\mathcal{H}} x ^h = {{x (1 - x ^{\mathcal{H}})}\over{1-x}}$. For large ${\mathcal H}$ we can approximate $N_{bu}$ by  $|\mathcal{D}_0| C _r {{q r^{\mathcal{H}-1}}\over{1 - q/r}}$ (because $(q/r)^{\mathcal{H}}$ will be small).

We calculate $N_{m_s} = q^{|\mathcal{H}|}|\mathcal{D}_0|$ for the complexity of model selection (which only requires thresholding at every point on the top-level lattice).

The complexity $N_{td}$ of the top-down pass is computed from equation~(\ref{eq:top_down}). At each level there are $r^{|\mathcal{H}|-h}$ nodes and we must compute $C_r$ computations for each. This yields complexity of $\sum _{h=1}^{|\mathcal{H}|} C_r r ^{|\mathcal{H}|-h}$ for each possible root node. There are at most $q^{|\mathcal{H}|} |\mathcal{D}_0|$ possible root nodes (depending on the results of the model selection stage). This yields an upper bound:
\begin{equation} N_{td} \leq |\mathcal{D}_0| C_r q^{|\mathcal{H}|} {{r ^{|\mathcal{H}|-1}}\over{1- 1/r}} \{1 - {{1}\over{r^{|\mathcal{H}-1|}}}\}.\label{eq:t_d}\end{equation}

Clearly the complexity is dominated by the complexity $N_{bu}$ of the bottom-up pass. For simplicity, we will bound/approximate this by:
\begin{equation} N_{s_o}= |\mathcal{D}_0| C _r {{q r^{\mathcal{H}-1}}\over{1 - q/r}}\label{eq:single_complexity}\end{equation}

Now suppose we perform inference for multiple objects simultaneously without exploiting shared parts. In this case the complexity will scale linearly with the number $|\mathcal{M}_{\mathcal{H}}|$ of objects. This gives us complexity:
\begin{equation} N_{m_o}= |\mathcal{M}_{\mathcal{H}}| |\mathcal{D}_0| C _r {{q r^{\mathcal{H}-1}}\over{1 - q/r}}\label{eq:multiple_complexity}\end{equation}

\subsection{Computation with Shared Parts in Series and in Parallel}

This section computes the complexity using part sharing. Firstly, for the standard serial implementation of part sharing. Secondly, for the parallel implementation.

Now suppose we perform inference on many objects with part sharing using a serial computer. This requires performing computations over the part-subpart compositions between elements of the dictionaries. At level $h$ there are $|\mathcal{M}_h|$ dictionary elements. Each can take $|\mathcal{D}_h|= q^h |\mathcal{D}|$ possible states. The bottom-up pass requires performing $C_r$ computations for each of them. This gives a total of $\sum _{h=1}^{\mathcal{H}} |\mathcal{M}_h| C _r |\mathcal{D}_0| q ^h = |\mathcal{D}_0| C_r \sum _{h=1}^{\mathcal{H}} |\mathcal{M}_h| q ^h$ computations for the bottom-up process. The complexity of model selection is $|\mathcal{D}_0| q ^{\mathcal{H}} \times (\mathcal{H}+1)$ (this is between all the objects, and the background model, at all points on the top lattice). As in the previous section, the complexity of the top-down process is less than the complexity of the bottom-up process. Hence the complexity for multiple objects using part sharing is given by:
\begin{equation} N_{p_s} = |\mathcal{D}_0| C_r \sum _{h=1}^{\mathcal{H}} |\mathcal{M}_h| q ^h.\label{eq:Mshared}\end{equation}

Next consider the parallel implementation. In this case almost all of the computations are performed in parallel and so the complexity is now expressed in terms of the number of "neurons" required to encode the dictionaries, see figure~(\ref{fig:comp5}). This is specified by the total number of dictionary elements multiplied by the number of spatial copies of them:
\begin{equation} N_{n} = \sum _{h=1}^{\mathcal{H}} |\mathcal{M}_h| q ^h |\mathcal{D}_0|.\label{eq:neurons}\end{equation}

The computation, both the forward and backward passes of dynamic programming, are linear in the number $\mathcal{H}$ of levels. We only need to perform the computations illustrated in figure~(\ref{fig:neural}) between all adjacent levels. 

Hence the parallel implementation gives speed which is linear in $\mathcal{H}$ at the cost of a possibly large number $N_n$ of "neurons" and connections between them.

\subsection{Advantages of Part Sharing in Different Regimes \label{sec:regimes}}

The advantages of part-sharing depend on how the number of parts $|\mathcal{M}_h|$ scales with the level $h$ of the hierarchy. In this section we consider three different regimes: (I) The \emph{exponential growth regime} where the size of the dictionaries increases exponentially with the level $h$. (II) The \emph{empirical growth regime} where we use the size of the dictionaries found experimentally by compositional learning \cite{LZhu10unsupervised}. (III) The \emph{exponential decrease regime} where the size of the dictionaries decreases exponentially with level $h$. For all these regimes we compare the advantages of the serial and parallel implementations using part sharing by comparison to the complexity results without sharing.

Exponential growth of dictionaries is a natural regime to consider. It occurs when subparts are allowed to combine with all other subparts (or a large fraction of them) which means that the number of part-subpart compositions is polynomial in the number of subparts. This gives exponential growth in the size of the dictionaries if it occurs at different levels (e.g., consider the enormous number of objects that can be built using lego).

\begin{figure}[tp]
\centering
\subfigure[]{
   \includegraphics[width=9.8pc] {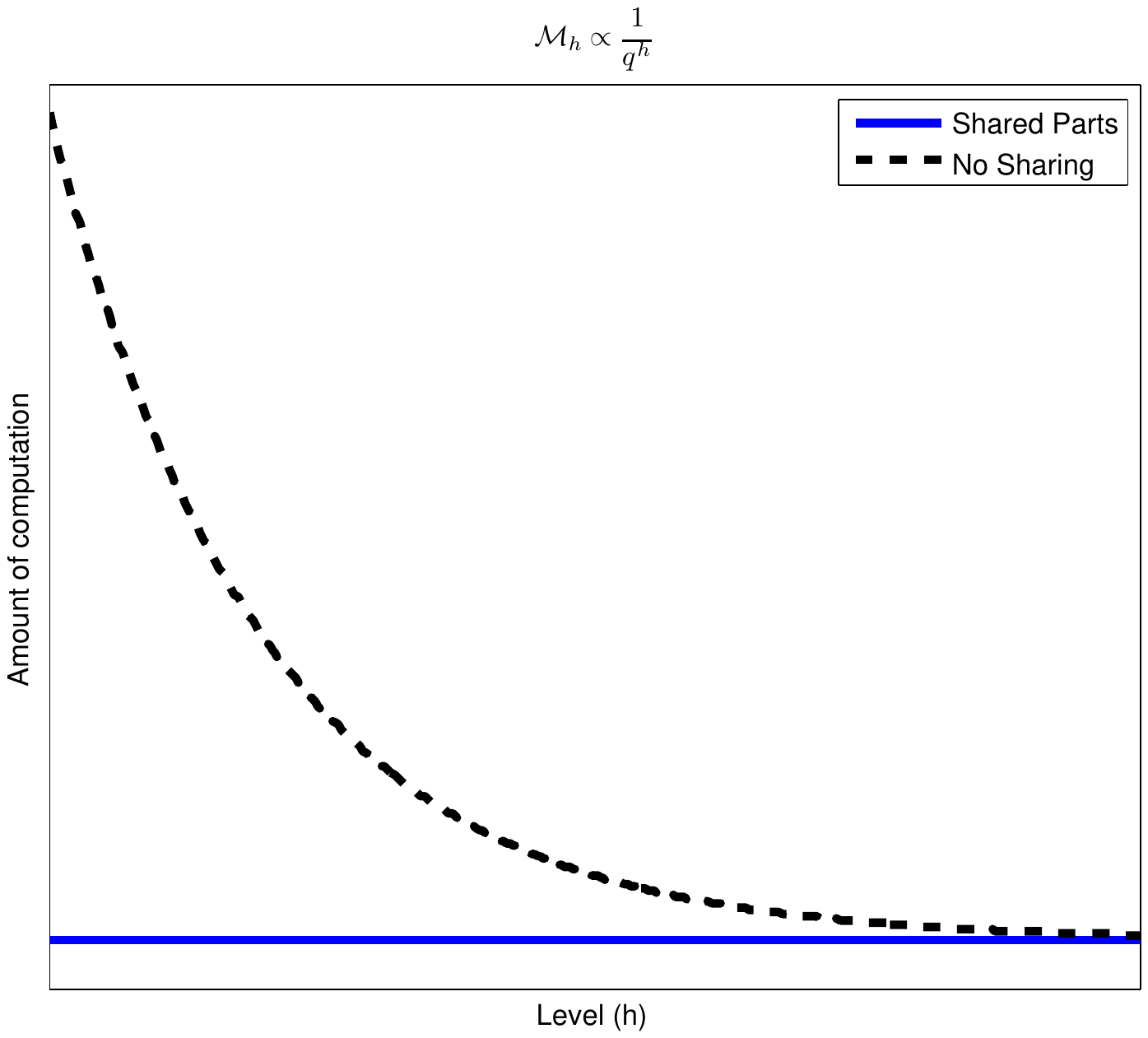}
   \label{fig:a}
 }
 \subfigure[]{
   \includegraphics[width=10.9pc] {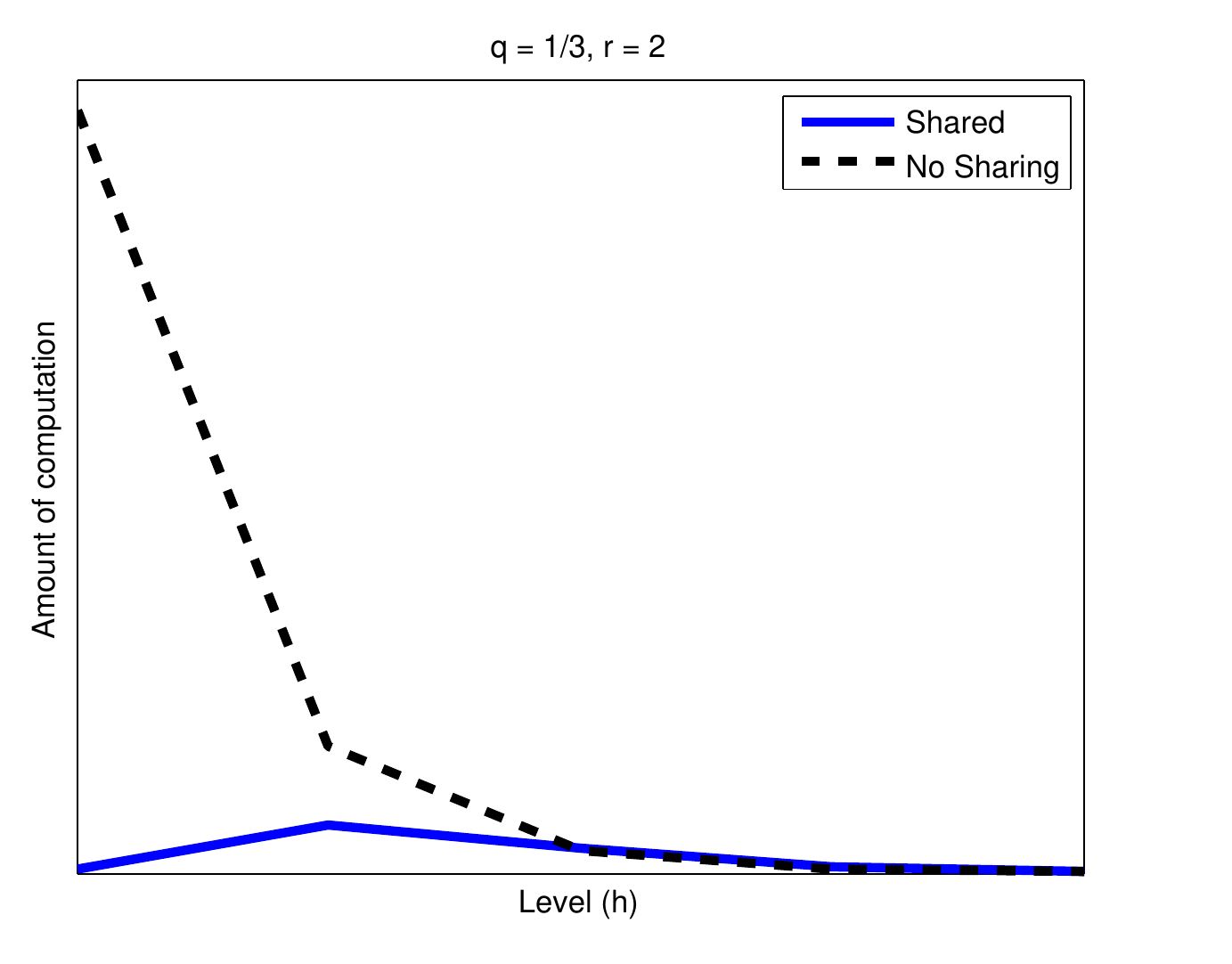}
   \label{fig:c}
 }
 \subfigure[]{   
   \includegraphics[width=10.2pc] {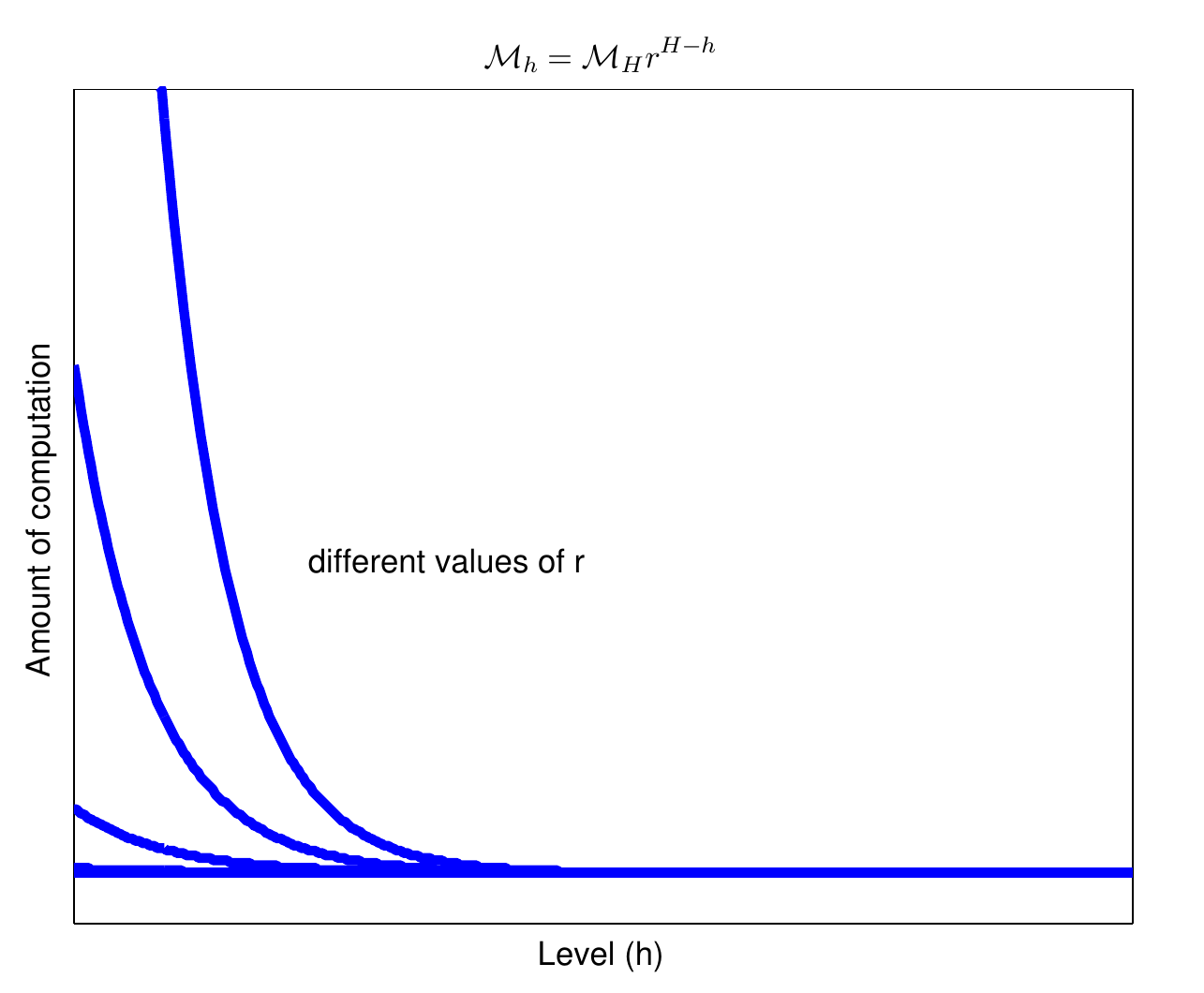}
   \label{fig:b}
 }
\caption{The curves are plotted as a function of h. Left panel:
The first plot is the case where $M_h = a / (q^h)$. So we have a constant cost for the computations, when we have shared parts. Center panel: This plot is based on the experiment of \cite{LZhu10unsupervised}. Right panel: The third plot is the case where $M_h$ decreases exponentially. The amount of computation is the same for the shared and non-shared cases. A set of plots with different values of $r$. \label{fig:roozbeh}} 
\end{figure}

An interesting special case of the exponential growth regime is when $|\mathcal{M}_h|$ scales like $1/q^{h}$, see figure~(\ref{fig:roozbeh})(left panel). In this case the complexity of computation for serial part-sharing, and the number of neurons required for parallel implementation, scales only with the number of levels $\mathcal{H}$. This follows from equations~(\ref{eq:Mshared},\ref{eq:neurons}). But nevertheless the number of objects that can be detected scales exponentially as $q^{\mathcal{M}}$. By contrast, the complexity of inference without part-sharing scales exponentially with $q$, see equation~(\ref{eq:multiple_complexity}, because we have to perform a fixed number of computations, given by equation~(\ref{eq:single_complexity}), for each of an exponential number of objects. This is summarized by the following result.
 
\emph{Result 1}: If the number of shared parts scales exponentially by $|\mathcal{M}_h| \propto {{1}\over{q ^h}}$ then we can perform inference for order $q^{\mathcal{H}}$ objects using part sharing in time linear in $\mathcal{H}$, or with a number of neurons linear in $\mathcal{H}$ for parallel implementation. By contrast, inference without part-sharing requires exponential complexity. 

To what extent is exponential growth a reasonable assumption for real world objects? This motivates us to study the empirical growth regime using the dictionaries obtained by the compositional learning experiments reported in \cite{LZhu10unsupervised}. In these experiments, the size of the dictionaries increased rapidly at the lower levels (i.e. small $h$) and then decreased at higher levels (roughly consistent with the findings of psychophysical studies -- Biederman, personal communication). For these "empirical dictionaries" we plot the growth, and the number of computations at each level of the hierarchy, in figure~(\ref{fig:roozbeh})(center panel). This shows complexity which roughly agrees with the exponential growth model. This can be summarized by the following result:
 
\emph{Result 2}: If $|\mathcal{M}_h|$ grows slower than $1/q^h$ and if $|\mathcal{M}_h| < r^{\mathcal{H}-h}$ then there are gains due to part sharing using serial and parallel computers. This is illustrated in figure~(\ref{fig:roozbeh})(center panel) based on the dictionaries found by unsupervised computational learning \cite{LZhu10unsupervised}. In parallel implementations, computation is linear in $\mathcal{H}$ while requiring a limited number of nodes ("neurons").

Finally we consider the exponential decrease regime. To motivate this regime, suppose that the dictionaries are used to model image appearance, by contrast to the dictionaries based on geometrical features such as bars and oriented edges (as used in \cite{LZhu10unsupervised}). It is reasonable to assume that there are a large number of low-level dictionaries used to model the enormous variety of local intensity patterns. The number of higher-level dictionaries can decrease because they can be used to capture a cruder summary description of a larger image region, which is another instance of the executive summary principle. For example, the low-level dictionaries could be used to provide detailed modeling of the local appearance of a cat, or some other animal, while the higher-level dictionaries could give simpler descriptions like "cat-fur" or "dog-fur" or simply "fur". In this case, it is plausible that the size of the dictionaries decreases exponentially with the level $h$. The results for this case emphasize the advantages of parallel computing.

\emph{Result 3}: If $|\mathcal{M}_{h}| = r^{\mathcal{H}-h}$ then there is no gain for part sharing if serial computers are used, see figure~(\ref{fig:roozbeh})(right panel). Parallel implementations can do inference in time which is linear in $\mathcal{H}$ but require an exponential number of nodes ("neurons").

Result 3 may appear negative at first glance even for the parallel version since it requires an exponentially large number of neurons required to encode the lower level dictionaries. But it may relate to one of the more surprising facts about the visual cortex in monkeys and humans -- namely that the first two visual areas, V1 and V2, where low-level dictionaries would be implemented are enormous compared to the higher levels such as IT where object detection takes places. Current models of V1 and V2 mostly relegate it to being a large filter bank which seems paradoxical considering their size. For example, as one theorist \cite{Lennie98} has stated when reviewing the functions of V1 and V2 ``perhaps the most troublesome objection to the picture I have delivered is that an enormous amount of cortex is used to achieve remarkably little''. Our complexity studies suggest a reason why these visual areas may be so large if they are used to encode dictionaries\footnote{Of course, this is extremely conjectural.}.

\section{Discussion \label{sec:discussion}}

This paper provides  a complexity analysis of what is arguably one of the most fundamental problem of visions -- how, a biological or artificial vision system could rapidly detect and recognize an enormous number of different objects. We focus on a class of hierarchical compositional models \cite{zhu08eccv,LZhu10unsupervised} whose formulation makes it possible to perform this analysis. But we conjecture that similar results will apply to related hierarchical models of vision (e.g., those cited in the introduction). 

Technically this paper has required us to re-formulate compositional models so that they can be defined on regular lattices (which makes them easier to compare to alternatives such as deep belief networks) and a novel parallel implementation. Hopefully the analysis has also clarified the use of part-sharing to perform exact inference even on highly complex models, which may not have been clear in the original publications. We note that the re-use of computations in this manner might relate to methods developed to speed up inference on graphical models, which gives an interesting direction to explore.

Finally, we note that the parallel inference algorithms used by this class of compositional models  have an interesting interpretation in terms of the bottom-up versus top-down debate concerning processing in the visual cortex \cite{DiCarlo:2012em}. The algorithms have rapid parallel inference, in time which is linear in the number of layers, and which rapidly estimates a coarse ``executive summary" interpretation of the image. The full interpretation of the image takes longer and requires a top-down pass where the high-level context is able to resolve ambiguities which occur at the lower levels. Of course, for some simple images the local evidence for the low level parts is sufficient to detect the parts in the bottom-up pass and so the top-down pass is not needed. But more generally, in the bottom-up pass the neurons are very active and represent a large number of possible hypotheses which are pruned out during the top-down pass using context, when ``high-level tells low-level to stop gossiping".

\section*{Acknowledgments} Many of the ideas in this paper were obtained by analyzing models developed by L. Zhu and Y. Chen in collaboration with the first author.  G. Papandreou gave very useful feedback on drafts of this work. D. Kersten gave patient feedback on speculations about how these ideas might relate to the brain. The WCU program at Korea University, under the supervision of S-W Lee, gave peace and time to develop these ideas.

\bibliographystyle{ieee}
\bibliography{papers2lib,papers2lib22}

\end{document}